# Computer-Aided Knee Joint Magnetic Resonance Image Segmentation - A Survey

Boyu Zhang[1], Yingtao Zhang[2], H. D. Cheng[2], Min Xian[3], Shan Gai[4], Olivia Cheng[5,6], Kuan Huang[1]

*Abstract*—Osteoarthritis (OA) is one of the major health issues among the elderly population. MRI is the most popular technology to observe and evaluate the progress of OA course. However, the extreme labor cost of MRI analysis makes the process inefficient and expensive. Also, due to human error and subjective nature, the inter- and intra-observer variability is rather high. Computer-aided knee MRI segmentation is currently an active research field because it can alleviate doctors and radiologists from the time consuming and tedious job, and improve the diagnosis performance which has immense potential for both clinic and scientific research. In the past decades, researchers have investigated automatic/semi-automatic knee MRI segmentation methods extensively. However, to the best of our knowledge, there is no comprehensive survey paper in this field yet. In this survey paper, we classify the existing methods by their principles and discuss the current research status and point out the future research trend in-depth.

*Index Terms*— Magnetic resonance imaging (MRI); Knee Osteoarthritis; Segmentation; Computer-aided Diagnosis (CAD); Bone.

## I. INTRODUCTION

Knee osteoarthritis (OA) is chronic and progressive disease characterized by structural changes in cartilage, bone, synovium, and other joint structures [1-3]. Knee OA is one of the leading cause of disability in the United States, which results a decrease in quality of life as well as a large financial burden on health care systems and society. Knee OA is prevalent and costly. It was estimated that symptomatic knee OA afflicts more than 9.9 million US adults in 2010 [4]. In the meantime, the estimated average lifetime treatment cost for each person diagnosed with knee OA was $140,300 [5].

There is growing understanding about knee OA after decades of research, although the pathogenesis of knee OA and the development mechanism is still unknown. One of the main effects of knee OA is the degradation of the articular cartilage, causing pain and loss of joints mobility [6]. Therefore, the volume and thickness of knee joint cartilage are the most important criteria to evaluate the course of Knee OA and to plan the treatments.

Different medical imaging modalities could be employed to investigate and quantitatively measure the knee joint cartilage, such as X-ray, Magnetic resonance imaging (MRI), and Computed Tomography (CT) [7]. Among them, X-ray and CT have similar responses on soft tissues and cartilage, and only can estimate cartilage thickness through the distance between bones [8]. A technology of CT, referred as knee arthrography, could overcome the above disadvantage, and display the articular cartilage clearly [9]. However, the pre-injection, which is requisite in knee arthrography, may cause pain and anxiety for some people besides possible complications [10]. MRI is the leading imaging modality for noninvasive assessment of the articular cartilage, and the cartilage deterioration can be analyzed effectively [11, 12].

MRI employs powerful magnetic field and radio frequency pulses to produce images of organs, soft tissues, bones, and virtually all internal body structures [13]. Knee MRI provides high-resolution images of the structures within the knee joint, including bones, cartilages, tendons, ligaments, muscles, and blood vessels, from different angles.

A regular knee MRI image may contain hundreds of slices (e.g., 160 slices per scan in OAI dataset [14-16]), depending on the sample rate. It costs hours for experienced radiologists to analyze a single scan [14]. The significant labor cost makes the diagnosis expensive, inefficient, and difficult to reproduce. The computer-aided segmentation technologies of knee MRI are in urgent need [17].

Automatic/semi-automatic segmentation is difficult due to the low contrast, noise, and especially the thin structure of cartilage. The efforts against such issues have been devoted for decades [11, 12, 17, 18]. Automatic volume and thickness measurements based on the manual label is the first step toward automatic analysis. The semi-automatic segmentation methods perform knee MRI segmentation with the intervention of doctors/radiologists through human-computer interaction. Semi-automatic segmentation could be accomplished using a variety of algorithms, such as active contours [19-22], region growing [23-25], Ray Casting [26, 27], Live Wire [28], Watershed [29], etc. Instead of sketching accurate boundaries manually, semi-automatic methods can generate accurate contours for each tissue in MRI scan based on a few landmarks, which is much more efficient and less afflictive. Semi-automatic methods have been widely applied to both scientific research and clinical practice because of the effectiveness.

[1]Department of Computer Science, Utah State University, Logan, Utah 84322, USA
[2]School of Computer Science and Technology, Harbin Institute of Technology, China
[3]Department of Computer Science, University of Idaho, Idaho Falls, Idaho 83402 USA
[4]School of Information Engineering, Nanchang Hangkong University, Nanchang 330063, China.
[5]Collingwood General and Marine Hospital, Collingwood, ON L9Y 3Z1, Canada
[6]Department of Surgery, McMaster University, Hamilton, ON L8S 4L8, Canada.

Accompany with the development of computer vision, pattern recognition, and machine learning, many automatic methods for knee MRI segmentation have been proposed, such as deformable models [30-32], classification-based methods [33-37], and graph-based methods [38-40], etc. Fully automatic segmentation is the major research goal in the field and has great potential.

Knee MRI segmentation, a specific subfield of multi-atlas image segmentation [41], has developed rapidly in recent decades. Researchers have conducted the surveys in related areas, such as brain MRI image segmentation [42], 3D medical image segmentation [43], Multi-atlas image segmentation [41], etc. However, to the best of our knowledge, there is no effective survey for knee MRI segmentation yet. This paper describes the knee MRI segmentation principles and gives a comprehensive review of computer-aided knee MRI segmentation methods. Moreover, we compare the existing methods advantages/disadvantages and point out the future research direction.

The rest of this paper is organized as follows: Section 2 summarizes the existing knee joint MRI segmentation methods; Section 3 describes the volume and thickness measurement methods; Section 4 gives a brief view of existing benchmarks and dataset as well as the performance evaluation metrics; and Section 5 concludes the entire paper and points out the future trend.

## II. KNEE MR IMAGE SEGMENTATION

### A. Overview

Accurate segmentation plays an essential role in computer-aided knee osteoarthritis diagnosis, and different methods were developed in recent decades. Depending on the level of automation, the existing methods could be categorized into manual segmentation, semi-automatic segmentation, and automatic segmentation. In the procedure of manual segmentation, the radiologists sketch the contours of different tissues on the MR images in a slice by slice manner. Manual segmentation is very labor intensive and time-consuming. It could take 3 to 4 hours for a well-trained radiologist to finish single scan. In addition, manual segmentation suffers from low reproducibility.

Semi-automatic methods are proposed to address the disadvantages of the manual method, by introducing computer algorithms. Semi-automatic methods, such as active contour [20-22, 44-51] and region growing [24, 25, 37, 52-55], provide a human-computer interface that users (radiologists/doctors) can input their professional knowledge (landmarks) into the system. Then the algorithm will generate the tissue boundaries based on the given landmarks automatically. Semi-automatic segmentation methods benefit from the professional knowledge and avoid intensive labor. Computer-Aided Diagnosis (CAD) systems equipped semi-automatic segmentation software are widely applied in the clinic because of the high accuracy.

The past decades have witnessed the rapid development of automatic knee joint segmentation. Researchers in both academia and medicine expect that by using computers with the related software, the systems can segment MRI images and quantize the cartilage thickness and volume with minimum or no human intervention. Fully automatic segmentation can be achieved by different approaches, such as pixel/voxel classification [33-37, 56-66], deformable model [20-22, 30-32, 44-51, 67-78] and graph-based methods [38-40, 79, 80]. Fully automatic knee joint segmentation suffers from the low contrast of MR image and the thin structure of the cartilage. Most state-of-the-art methods share the similar procedure. They start at the bone segmentation, which is less complex; and then sketch the boundaries of cartilage based on the result of bone segmentation.

The cartilage volume and thickness are the principal factors in OA diagnosis and progress evaluation. In general, the measurement of cartilage volume is quite straightforward. The task can be done by simply counting the number of pixels when the boundaries of cartilage are determined. However, the evaluation of thickness is more complex. Several studies define the thickness of 2-D MR image slice [21] while others define the thickness of 3-D space [48]. 3-D Euclidean Distance Transformation (EDT) [81] is commonly used to determine the distance between the cartilage surface and the Bone-Cartilage Interface (BCI). There are also researches transfer 3-D cartilage structures into thickness map and find local cartilage loss among the map [46].

Bones are the biggest and the most salient structure in the knee joint. Other structures, such as cartilage, are attached to the bones. The segmentation of bones is simpler than cartilage segmentation due to higher contrast and much clearer edges. Many methods perform cartilage segmentation based on bone segmentation results; while other methods solve both bone and cartilage segmentation at the same time. There are also researchers develop end-to-end systems to solve the knee MR image segmentation problem in full.

### B. Deformable Models

Deformable models [82], ones of the most intensively studied model-based approaches for computer-aided medical image analysis, can accommodate the often-significant variability of biological structures over time and across different individuals. This subsection surveys the applications of deformable models to knee joint MRI segmentation.

#### 1) Active Contours

Active contour models [20] [83] are very popular for semi-automatic medical image segmentation. An active contour model is defined by a set of points $v_i$, where $i = 0, 1, \ldots, n-1$; the internal elastic energy $E_i$ and external energy $E_e$. The internal energy controls the deformations made to the snake, and the external energy term drives the curve to the object contour. The algorithm seeks the best segmentation by minimizing the energy function. A bunch of active contour models were introduced for knee MRI segmentation, such as B-spline [44, 48], Bezier spline [45], geodesic snake [50], GVF snake [51], and snake [22, 46, 49].

The performance of active contour models depends very much on the selection of initialization points. However, there existed hardly any literature about how to select the initial points in the field of knee MRI analysis. Existing methods [20-22, 44-51] started from the points inputted by human operators. Active contour models benefit from the participation of human experts, as well as the real-time guidance and correction. However, the dependence on human participation hindered the development of the algorithms.

Different models and energy functions were utilized in knee MRI segmentation. Carballido-Gamio et al. [45] proposed a Bezier spline-based method. The edges were found based on the first derivative of the brightness using bicubic interpolation along the line profiles. Cohen et al. [48] used an interpolated cubic B-spline curve to fit the initial set of points. The algorithm evaluated the image gradient vector using the Prewitt convolution kernel [84]. Parametric spline models were convenient for user interaction and were easier to implement. However, the global information was not considered in parametric spline models, and the models were difficult to converge.

The classic snake models considered both local features and global shapes. Kauffmann et al. [62] minimized the edge energy to deform the contour over the most significant neighboring image edge. A symmetric Windowed-Sinc filter was utilized to conduct contour regulation. Lynch et al. [44, 85] considered both the sign of the edge strength and the local direction of the boundary in the classical snake model. Brem et al. [49] and Duryea et al. [22] utilized snake model for knee MRI segmentation. The classical snake models were sensitive to the initialization. The optimization of these models was non-convex, and it could fall into a local minimum. Another disadvantage of the classic snake models was their difficulties in fitting concave boundaries.

By introducing gradient vector flow as a new external force, GVF snake model could solve the sensitivity to initialization and the fitting problem on concave shape boundary. Tang et al. [51] utilized GVF snake for knee MRI segmentation, and the experimental results showed that the method achieved better reproducibility, comparing to previous studies. Lorigo et al. [50] utilized GAC snake model for knee MRI segmentation and achieved accurate results. GAC snake model was based on the relationship between active contours and the computation of geodesics or minimal distance curves. The algorithm was robust to initialization and could handle the topological variances of the curve. However, the algorithm had difficulties when the edges were blur or with low contrast.

Methods based on active contour model segmented knee MRI in a slice-by-slice manner. The broad application of slice selection technics could speed up the segmentation process, reduce the labor cost, and improve the performance. The method in [22] performed a two-dimensional segmentation on each slice of the MRI. A slice near the center of the cartilage plate was selected, and a seed point on the bone-cartilage margin was marked. The software then employed an automated edge tracking algorithm to segment the cartilage on the slice. An automatic active contour algorithm was utilized to refine the segmentation. Once a central slice was segmented, the software proceeded to an adjacent slice using the computer-delineated margins from the previous slice and an active contour edge-detection algorithm conducted automatic segmentation. A similar strategy was used in [44] and [49].

Active contours had been widely applied to knee MRI segmentation. High precision could be achieved by making a minimal number of user inputs. Active contours also allowed an experienced user to guide, correct and validate the segmentation results in real-time. However, the active contours had limitations in convergence, and the optimization problem leads uncertainty and poor stability of the segmentation. Table 1 presents the existing knee MRI segmentation methods based on active contour model and lists the weakness of the methods.

*Table 1 Active Contour for Knee Joint Segmentation*

| Method | Publication | Target | Weakness |
| --- | --- | --- | --- |
| Geodesic Active Contour. | Lorigo et al. [50]. | Bone | GAC snake model was sensitive to edge clarity. |
| Cubic B-spline; Bezier Spline. | Cohen et al. [48]; Lynch et al. [44]; Carballido-Gamio et al. [45, 47]. | Cartilage | Further process was needed for these methods; The methods were sensitive to initialization; The optimization was non-convex; The convergence was not guaranteed. |
| Directional GVF Snake. | Tang et al. [51]. | Cartilage | GVF snake model could not handle long, thin concave shape. |
| Active Contour. | Duryea et al. [22]; Kauffmann et al. [46]; Brem et al. [49]. | Cartilage | Sensitive to initialization; The optimization is non-convex, and the algorithm will fall into local minimum easily; |

*2) Statistical Shape Models*

The Statistic Shape Models (SSMs) [43] represent the shape of target objects using a set of n landmarks and learn the valid ranges of shape variation from a training set of the labelled images. The shape model can be matched with new image using different algorithms, where the Active Shape model (ASM) [69, 86] and Active Appearance Model (AAM) are the most widely studied. Compared to the active contour models, AAMs are more informative and allow the algorithms to locate target object without operator intervention. They could converge to object boundaries base on shape restrictions, where the active contour models are powerless.

Both AAM and ASM use the same underlying statistical model of the target object shape. However, ASM seeks to match a set of model points to an image and AAM seeks to match both the position of the model points and a representation of the object texture to an image. There are two key differences between the two algorithms [87]: First, ASM only uses the image texture in small regions around the landmark points, whereas AAM uses the appearance of the whole region; second, ASM minimize the distance between model points and the corresponding points found in the image, whereas AAM seeks to minimize the difference between the synthesized model image and the target image. In general, ASM is faster, and it locates feature point location more accurately, comparing to AAM. However, AAM gives a better match to the texture.

Solloway et al. [21] used 2D ASM to segment bone and cartilage in a slice-by-slice manner. In addition to modeling shape variation, the method modeled the gray-level appearance of the objects of interest by examining small image patches around each landmark point. They claimed that their local appearance model could reduce the dependency of sharp edges. 2D methods could hardly model 3D structures because the relationship between slices was ignored. Moreover, the results of 2D methods need further processing in most cases. 3D methods attract more attention comparing to 2D methods.

ASM and AAM are built utilizing labeled MRI samples. Because of the inter-object variations, finding the correspondences between the dense landmarks is a crucial step for both model construction and segmentation. Most studies [30-32, 67, 70, 71, 76, 78] utilized minimum description length (MDL) [88] and its variations [89] to find the correspondences. Besides MDL, Schmid et al. [77] preferred the approach of Dalal et al. [90]. They claimed their method did not require any specific shape topology and was appropriate to refine a previous correspondence.

Registration technologies are widely applied to model initialization. Fripp et al. [67] argued that the method in [21] was sensitive to initialization. They introduced a robust affine registration to initialize the model automatically. In [30, 31], the authors introduced a robust affine registration to initialize the model automatically. The introduction of registration could enhance the robustness to initialization.

As an essential component of ASM and AAM algorithms, principal component analysis (PCA) was applied in most studies to describe the main directions of shape variation in a training set of example shapes. Beside PCA which captures global shape variations, Markov Random Fields (MRF) were used to capture global shape variations and local deformations, respectively [74].

*Table 2 The ASMs and the AAMs for Knee MRI Segmentation*

|  | Model | Targets | Weakness |
|---|---|---|---|
| Solloway et al. [21]. | ASM (2D) | Bone/Cartilage | 1. 3D structures were ignored in the model; 2. Post processing was needed. |
| Fripp et al. [30, 31, 67]; Schmid et al. [74, 77]. | ASM (3D) | Bone | 1. The search for initial model pose parameters can be very time consuming; 2. The initialization was based on manually defined landmarks. |
| Gilles et al. [75]; Seim et al. [73]. | ASM (3D) | Bone/Cartilage | 1. Such models only reach a local optimum and depend heavily on their initial position; 2. The cartilage segmentation largely depends on the preset parameter 'thickness.' |
| Vincent et al. [32]; Williams et al. [70, 71]. | AAM | Bone/Cartilage | Variation outside these spaces cannot be properly captured if no subsequent relaxation step is used; |

For ASM, iterative approaches are usually employed to find the best match with new images. For each iteration, the algorithm examines the region around current model points $(X_i, Y_i)$ to find the best nearby match $(X'_i, Y'_i)$, then updates the parameters $(t_x, t_y, s, \theta, b)$ for the best fit of the model to the newly found point set X'. The process will repeat until convergence. Instead of searching around each model point locally, AAMs seek to minimize the difference between a new image and a synthesized appearance model. Under the restriction of a set of model parameters, a hypothesis for shape $x$, and texture $g_m$ is generated for the instance. Then the difference between the hypothesis and the image is evaluated by computing $\delta_g = g_s - g_m$, where $g_s$ is the image texture. The final segmentation is done by minimizing the magnitude of $|\delta_g|$.

ASM and AAM can model the structure in the knee joint well and could be applied to both bone and cartilage segmentation. These models perform better on bone segmentation than on cartilage segmentation because of the thin structure of the cartilage. AAMs and ASMs highly rely on the datasets used and thus are not easy to reproduce. Table 2 summarizes the existing studies based on AAMS and ASMs.

*C. Classification-based Methods*

Both voxel classification-based methods and region-growing methods share the same principle that one or more classifiers are learned from training samples to distinguish foreground from background. Classification based methods are robust and could benefit from the growing size of the training data. Furthermore, voxel classification-based methods are the most effective and straightforward way to achieve fully-automatic knee MRI segmentation.

*1) Voxel Classification*

A lot of efforts had been devoted to voxel classification (VC) based methods in recent years. These methods have obvious advantages. First, the philosophical model is concise, and the computing process is straightforward; second, a variety of features, both local and global, could be combined to achieve a better performance; and third, these methods could obtain better performance by effectively utilizing substantial training samples. The over-segmentation of the cartilage surface is one of the main drawbacks. Moreover, the VC based methods are inefficient compared with other methods because of the considerable number of voxels.

The first study on VC based knee MRI segmentation, which was also considered as the first fully-automatic knee joint MRI segmentation [37], was reported by Folkesson et al. [33]. The framework included a two-class kNN classifier for distinguishing foreground (articular cartilage) from other tissues, and a three-class kNN classifier assigned each voxel as bones (tibial and femoral), cartilage and background. A variety of features were used that included the 3-jet (first, second and third order derivatives on $x, y, z$ [91]), the Hessian matrix, the structure tensor (ST) [92], etc. They claimed that Hessian and ST were the most effective features.

The above method was extended in [58], where the combination of two binary classifiers replaced the single three-class classifier. It was reported that the combination of binary classifiers outperformed the three-class classifier. The method achieved 84.14%, 99.89%, 0.811 on sensitivity, specificity, and DSC, respectively. A further study improved the efficiency by focusing mainly on the cartilage voxels [64].

The 2.5 hour segmentation duration of the original algorithm impeded its application [33]. In recent years, more VC based methods were proposed to accompany with the rapid advance of the computational power. A two-stage classification framework was utilized to achieve accurate cartilage segmentation [57]. They used kNN as the first classifier for highly accurate background detection. Then they use a non-linear SVM to make the final decision on voxels that were not identified.

Shan et al. [34] developed the algorithm based on Folkesson's methods [33, 58, 64] by using a probabilistic version of kNN classifier to integrate the classification results into a Bayesian framework. The spotlight of this work is to introduce multi-atlas registration for generating the spatial prior. An extensive validation of the proposed method and the results were summarized in [35]. Moreover, this method was utilized in [60] as the foundation for disease region detection using knee MRI.

Wang et al. [63] presented a novel multiresolution patch-based segmentation framework that established a coarse initial segmentation quickly using the lowest resolution image and then refined the results using higher resolutions subsequently. kNN was utilized as the classifier and slices were selected based on a histogram of gradients.

*Table 3 VC based Methods for Knee Joint Segmentation*

|  | Classifier | Pros | Cons |
|---|---|---|---|
| Folkesson et al. [33] | KNN | The method could segment knee MRI fully automated. | 1. Very low computational efficiency; 2. Over-segmentation on the BCI; 3. The training process is time-consuming; 4. A large number of training samples are needed. |
| Folkesson et al. [58] | KNN | The combination of two-class KNN outperformed the three-class KNN [33]. | |
| Prasson et al. [57] | KNN+SVM | The computational efficiency was improved by using two-stage classification framework. | |
| Shan et al. [34] | Probabilistic KNN | The spatial prior generated by multi-atlas registration improved both performance and computational efficiency. | |
| Wang et al. [63] | KNN | The introduced novel multiresolution patch-based segmentation framework allowed a coarse-to-fine segmentation. | |
| Liu et al. [62] | Random Forest | Improved the performance by introducing context information. | |
| Dam et al. [56] | KNN | The rigid multi-atlas registration allowed the multi-structure segmentation. | |
| Prasson et al. [59] | CNN | The introduced deep feature improved the performance. | |

The voxel classifier focuses on the local feature of each voxel and its position. To improve the performance, researchers try to add the global feature into the voxel classification. Liu et al. [62] utilized random forest as voxel classifier and combined appearance features and context feature. They claimed that the cartilage was typically small and thus the tentative segmentation results were too unreliable to provide satisfactory context information. They introduced a multi-atlas registration process to solve the problem.

Dam et al. [56] proposed a segmentation framework combining rigid multi-atlas registration with voxel classification in a multi-structure setting. They registered two given scans by maximizing the normalized mutual information [93] (NMI) using L-BFGS-B (Fortran subroutines for large-scale bound constrained optimization) optimization [94]. They solved the unbalance between different structures by utilizing a simple sampling scheme. In the training process, kNN was selected as a voxel classifier and floating forward feature selection [95] was used for feature selection based on the Dice volume overlap score. The experiments utilized 1907 samples.

A basic idea to improve the performance of voxel classification methods is to improve the classifier. Prasson et al. [59] introduced the deep feature to knee MRI segmentation. They used a Convolutional Neural Network as the classifier, and the features were similar to that in [64].

As the most popular methods, voxel classification-based methods consider not only the local appearance but also the relationship among slices. The current trend is to combine the global feature with voxel classification method. How to reduce the dependency on position information is another important problem to solve. Table 3 summarizes the representative VC based segmentation algorithms.

*2) Region Growing*

Region growing [23] is one of the most popular image segmentation methods. As the first step of region growing, one or more seed points must be selected. Then the regions grew from the seed points to the adjacent points depending on a region membership criterion. The criterion could be, for instance, pixel intensity, grayscale texture, color, etc. Region growing methods are categorized as classification-based methods since the above process could be solved as a binary classification problem.

Region growing methods have been extensively studied in other areas as well. In knee MRI segmentation, most region growing based methods are semi-automatic because the seed points need to be selected manually [25]. An automatic method to select seed points was presented [37]. In this method, a candidate was generated with a 3-D Gaussian, and a classifier determined whether it was cartilage. The candidates that were classified into cartilage were selected as the seeds. Then the result of the voxel classification was used as the criteria of region growing.

Region growing methods have difficulties in handling the thin structure of cartilage; therefore, it is widely applied on bone segmentation, and only few researchers utilize it on cartilage segmentation.

*D. Graph-based Methods*

Graph-based segmentation is widely used in multi-surface and multi-object segmentation tasks. The methods model the initial segmentation by utilizing graph model and then optimize the model by minimizing specific cost functions. Most graph-based methods need pre-segmentation results as the initialization. There are methods constructed pre-segmentation results automatically [37, 53].

A layered optimal graph image segmentation of multiple objects and surfaces (LOGISMOS) was proposed [40]. The algorithm took a pre-segmentation result as the input, then performed the segmentation task by utilizing graph model. The final segmentation was accomplished by optimizing the cost function.

In [39], a level set based algorithm for automated bone surface pre-segmentation was used as initialization; then the implicit surface was converted to an explicit triangular mesh which was optimized. Finally, the mesh was used to initialize a graph in a narrow band around the pre-segmented bone surface, and a multi-surface graph search algorithm was used to obtain the precise cartilage and bone surfaces simultaneously.

In [38], Li et al. employed a graph-construction scheme based on triangulated surface meshes obtained from a topological pre-segmentation and utilized an efficient graph-cut algorithm for global optimality.

In [80], seeds were manually marked to indicate cartilage and non-cartilage. The seeds were propagated slice-by-slice, and a graph cuts algorithm segmented the cartilage regions automatically. The optimal segmentation satisfies the global criteria.

The pre-segmentation could be performed automatically. In [79], a content-based, two-pass disjoint block discovery mechanism was designed to support automation, segmentation initialization, and post-processing. The block discovery was achieved by classifying the image content to bone and background blocks according to their similarity to the categories in the training data collected from typical bone structures and background. The classified blocks are used to design an efficient graph-cut based segmentation algorithm. This algorithm required constructing a graph using image pixel data followed by applying a maximum-flow algorithm which generates a minimum graph-cut as an initial image segmentation. Content-based refinements and morphological operations were applied to obtain the final segmentation.

Graph-based methods achieve impressive results on knee MRI segmentation as widely applied models. The efficiency of graph-based methods is lower than that of deformable models but higher than that of VC-based methods. One of the major disadvantages of graph-based methods is that these methods need pre-segmentation results to initialize. Graph-based methods work well for refining the segmentation, but it is difficult to build an end-to-end system.

### E. Multi-atlas Segmentation

Graph-based methods achieve impressive results on knee MRI segmentation as widely applied models. The efficiency of graph-based methods is lower than that of deformable models but higher than that of VC-based methods. One of the major disadvantages of graph-based methods is that these methods need pre-segmentation results to initialize. Graph-based methods work well for refining the segmentation, but it is difficult to build an end-to-end system.

The segmentation of Knee MRI is a typical Multi-Atlas Segmentation (MAS) [41] problem. MAS problems could be solved by propagating the label from one labeled slice to other slices. This process can also be understood as registering unlabeled slices to one or more labeled slices. Atlas-based methods [85, 96-98] had been studied extensively since the last century.

In [98], segmentation was achieved through registration which aims at reforming the atlas such that the conditional posterior of the learned (atlas) density was maximized. Authors also employed primal/dual linear programming to accelerate the registration process.

In some research, registration is used to provide the initial segmentation. Lee et al. [96] registered all training cases to a target image by a nonrigid registration scheme and selected the best-matched atlases. A locally weighted vote (LWV) algorithm was applied to merge the information from the atlases and generate the initial segmentation.

Atlas-based methods could combine with other methods based on classification and statistical model. Lynch et al. [46] utilized 3D registration technology to remove human interaction and subjectivity from the process.

Atlas-based segmentation is one of the most widely applied methods for MAS. The idea is easy to understand, and there are many methods. However, the performance of registration process is time-consuming, and at least one labeled slice is needed.

### F. Summary

Each knee MRI segmentation method has its advantages and disadvantages. The ACs start at the initial points provided by a human expert. The Region Growing algorithms need pre-selected seed points. Graph-based methods need initialization. The atlas-based methods need at least one labeled slice as initialization. There are difficulties to provide initialization for the above methods. Comparatively, the fully automated framework could be developed more naturally utilize SSMs, AAMs, and classification-based methods.

Computational complexity is a major factor. The deformable models are the most efficient. The Atlas-based methods need higher computational power because of the global searching and optimization. The computational complexity is high for the VC-based classification methods. These methods need to process each voxel in the feature space. The graph-based methods need both high storage and high time complexity. Providing a proper initialization could accelerate these algorithms.

*Table 4 Comparison of Different Methods*

| Method | Pros | Cons |
|---|---|---|
| SSMs and AAMs | 1. The methods could be fully-automatic; 2. The models could handle incomplete boundaries well. | 1. The methods have difficulties for cartilage modeling; 2. The performance largely depends on the representativeness of the training samples. |
| Active Contour | 1. Low computational complexity; 2. The methods could achieve accurate result with expert interaction. | 1. It is hard to build automatic system based on Active Contour; 2. Convergence problem. |
| Voxel Classification | 1. Concise model; 2. High accuracy; 3. The performance could be improved by providing more training samples. | 1. Over-segmentation and the need of post-processing; 2. Computational complexity is high. |
| Graph-based | 1. High accuracy; 2. Global optimization could be achieved. | 1. High computational complexity and high storage requirement; 2. Initialization is needed in most cases. |
| Atlas-based | 1. High accuracy; 2. The process and the results are intuitionistic. | 1. Existing labeled slices should be provided; 2. Difficulties in handling the variations |

SSMs and AAMs could handle the missing boundaries well. ACs have problems in converging, and missing boundary may lead to unstable. In general, ACs are two-dimensional methods and SSMs and AAMs could work in both 2D and 3D. The 3D structure could provide more information and may produce better performance. SSMs and AAMs learn parameters from training samples. Thus, the sample selection plays a key role in the system.

Deformable models work well on bone segmentation. However, SSMs and AAMs are not suitable for cartilage segmentation well because of the thin and curve structure of the cartilage. ACs could be applied to cartilage segmentation directly with human intervention. SSMs and AAMs obtain better performance by finding initial points or correspondences based on the results of bone segmentation.

The presence of over segmented protrusions on the surfaces of the segmented cartilage compartments was the main drawback of the VC based methods, besides the high computational complexity. The cartilage edge is not smooth enough in the results of VC-based methods, and there are often holes in disconnected areas. However, classification-based methods develop rapidly. These methods could make structure information into segmentation and gain benefits from the growing size of the training data.

*Table 5 Typical Thickness Measurement Methods.*

| Authors | 2D/3D | Method | Weakness |
|---|---|---|---|
| Solloway et al. [21]. | 2D | 'M-Norm' | 1. The distance defined in 2D space could not accurately reflect the true thickness; 2. The method is unstable. |
| Tang et al. [51]. | 2D | 'T-norm'. | The distance defined in 2D space could not accurately reflect the true thickness. |
| Cohen et al. [48]. | 3D | Ray casting. | The method is unstable when the BCI is not smooth. |
| Stammberger et al. [81, 99]. Carballido-Gamio et al. [45, 47]. Fripp et al. [31]. | 3D | 3D Euclidean distance transformation. | High computational cost. |
| Williams et al. [100, 101]. | 3D | 'Spans and Ridges'. | The method is unstable when the BCI is not smooth. |
| Shan et al. [34, 35] Huang et al. [60]. | 3D | 3D Laplace-equation. | High computational cost. |
| Kauffmann et al. [46]. | 3D | Thickness map | High computational cost. |

III. QUANTITATIVE MEASUREMENT

The volume and thickness of the cartilage are crucial factors in OA diagnosis and progress evaluation. The measurement methods [12] play key roles in clinical trials as well as in scientific research. This section summarizes existing measurement methods for cartilage volume and thickness.

Literature shows that the volume measurement methods have been extensively investigated. By providing the inner and outer surface/boundary of the cartilage, the measurement of cartilage volume could be accomplished straightforwardly by counting pixels or voxels; therefore, it is considered as a solved problem in most studies [24, 28-31, 52].

The measurement of the thickness is more complex comparing to the volume measurement. The most widely accepted concept of thickness is the shortest distance between the base points on the BCI and their corresponding points. However, researchers employ different definitions of thickness in studies. The calculation of thickness could be performed in both 2D and 3D spaces, and the strategies to find the corresponding points can be different.

In some early studies [24] and [102], the thickness was defined as the distance between base points on BCI and the corresponding intersection points on the cartilage surface. Solloway et al. [21] proposed a 2D method referred as 'M-norm.' They calculated the medial axis between inner and outer surfaces and equally spaced a certain number of points along the medial axis. At each point, the normal was cast toward both inner and outer boundaries, and the thickness was saved as the distance between intersection points. Tang et al. [51] calculated the norm vector based on the plain composed of the point and its nearest points, instead of only one point. The so-called 'T-norm' method could avoid the possible instability.

In 3D spaces, the most widely used methods are 3D Euclidean distance transformation [31, 45, 66, 81, 99] and Laplace equation [31, 60]. Some heuristic searching methods are also used to find the nearest points, such as searching along the normal direction from the inner cartilage [103] and 'Spans and Ridge' [100, 101]. Some researchers transfer the original coordination system into a BCI based coordination system in which it can simply calculate the thickness map [46]. Table 5 summarizes the typical methods for thickness measurement.

IV. BENCHMARK AND EVALUATION

Use Effective dataset or benchmark is essential for doctor training and scientific research. Researchers in knee MR image segmentation area utilize different datasets to verify their methods. This section summarizes the experimental data and the current public benchmark and dataset.

Collecting experimental data for knee MR image segmentation research is difficult. A lot of studies were performed with a relatively small number of samples [25, 44, 45, 50, 51, 54, 55]. Lynch et al. [44] performed their experiments on a dataset with four patients (2 with OA and 2 with meniscal surgery). Kauffmann et al. verified their method on only 2 OA patients. There are also some studies on porcine knees [45], cadaveric human knee [24, 48, 51], or even synthetic MRI data [46].

The Osteoarthritis Initiative (OAI) [16] establishes and maintains a natural history database for osteoarthritis that includes clinical evaluation data, radiological (x-ray and magnetic resonance) images, and a biospecimen repository from 4796 men and women, ages 45-79, enrolled between February 2004 and May 2006. Currently, it provides up to 108-month follow-up of these cases. The data in OAI are widely utilized in different studies. In the viewpoint of knee MR image segmentation, the disadvantage of OAI data is that no segmentation results (ground truths) are provided. The MICCAI grand challenge published SKI10 data [14] with fully verified segmentation results. SKI10 data contains 50 testing data and 100 training data. In 2012, MICCA grand challenge published PROMISE12 dataset for knee MR image segmentation [104]. The dataset contains 49 training samples and 29 testing samples. All samples in PROMISE12 are segmented. The disadvantage of this dataset is the segmentation results are not validated. Table 4 presents the content and feature of existing public benchmarks for knee MR image segmentation research.

*Table 6 Benchmarks for Knee MR Image Segmentation*

| Dataset | Content | Ground Truth |
|---|---|---|
| OAI [16] | 4,796 participants as of Dec 2016, longest 108 months' follow-up | No segmentation results |
| SKI10 [14] | 100 training samples, 50 test samples | Segmented and validated |
| PROMISE12 [104] | 49 training samples, 29 testing samples | Segmented, but not validated |

Different objects and criteria could be utilized to evaluate the performance of knee MR image segmentation methods. Most studies employ the volume of cartilage and thickness of cartilage to evaluate the performance. For semi-automatic segmentation methods, the test-retest reproducibility is a principal factor for evaluation.

In general, the ground truths of knee MR image segmentation come from the results sketched by radiologists. The manual segmentation results are used as the ground truths for semi-automatic and automatic segmentation evaluation. The semi-automatic segmentation results by professional users are employed to evaluate the performance of automatic methods as well.

The most accepted criteria are Dice similarity coefficient (DSC), sensitivity and specificity. Let the segmentation result and the ground truth be A and B; the DSC value, sensitivity, and specificity could be calculated as follows:

$$DSC(A, B) = \frac{2|A \cap B|}{|A| + |B|}$$

$$Sensitivity(A, B) = \frac{|A \cap B|}{|B|}$$

$$Specificity(A, B) = \frac{|A^c \cap B^c|}{|B^c|}$$

Sensitivity and specificity measures are true classification ratios of cartilage and background classes. DSC is maximized in the case of a good compromise between sensitivity and specificity. Therefore, it is an important indicator for evaluating the accuracy of segmentation.

## V. Conclusion and Future Trend

Computer-aided knee MRI segmentation has immense potential in clinical diagnosis, as well as in scientific research. The central issue in knee MRI segmentation is to find the boundary of bone and cartilage accurately. For bone segmentation, the existing literature supports that the deformable model-based methods can handle bone segmentation well. The region-growing based semi-automatic methods also achieve satisfactory results. Voxel/pixel classification-based methods could handle bone segmentation well. However, researchers prefer deformable models for bone segmentation due to the higher computational efficiency. The segmentation of articular cartilage still needs further investigation, especially the fully-automatic cartilage segmentation. 2D deformable models, such as snake, can solve cartilage segmentation problem well in semi-automatic manner. But 3D deformable models are not robust due to the thin structure of cartilage. Voxel classification-based methods are growing rapidly, even with their high computational cost.

There are two strategies for knee MRI segmentation. The first strategy is to label one or a group of slices first and then propagate the labels to unlabeled slices. The other way is to work in 3D feature space. Current literature supports that the 3D feature space strategy performs better than label propagation by considering the relationship between slices.

OAI is the largest and the most important dataset for the knee OA research. However, the samples in OAI are not labeled for segmentation purpose. The sample numbers of the current public benchmarks are still not sufficient for the studies.

Large-scale benchmark with validated segmentation results (ground truths) is important and essential for future studies. Researchers need to validate their methods by using the benchmark including the labeled follow-up cases.

Accurate and efficient classifiers are in urgent need for cartilage segmentation research. The classifier performance is the key factor affecting the segmentation results. The pattern recognition and machine learning technologies provide more powerful classifiers such as CNN, FCN, etc. These methods have great potential to achieve better results.